\documentclass[letterpaper, 10 pt,conference]{ieeeconf}  
\usepackage{amsmath} 
\usepackage{amssymb} 

\newcommand{\proj}{\mathcal{P}}
\newcommand{\backproj}{\mathcal{Q}}
\usepackage{authblk}

\usepackage{ltl} 
\usepackage{pgfplots}
\usepackage[noadjust]{cite}

\usepackage[vlined]{algorithm2e}
\usepackage{algorithmic}
\usepackage{color}
\usepackage{subfig}
\usepackage{tikz}
\usepackage{hyperref}
\usetikzlibrary{arrows,fit,shapes,automata}
\usetikzlibrary{positioning,fit,calc,shapes}
\usetikzlibrary{decorations.fractals}
\usetikzlibrary{decorations.markings}

\newcommand{\Suda}[1]{{\textcolor{blue}{ \textbf{Suda:} #1 $\spadesuit$ }}}
\newcommand{\Louis}[1]{{\textcolor{red}{ \textbf{Louis:} #1 $\spadesuit$ }}}

\newtheorem{example}{Example}
\newtheorem{problem}{Problem}
\newtheorem{theorem}{Theorem}

\newcommand{\init}{\mathsf{init}}
\newcommand{\belief}{\mathsf{belief}}

\newcommand{\vis}{\mathit{vis}}
\newcommand{\succs}{\mathit{succ}}
\newcommand{\beliefs}{\mathcal{P}(L_t)}

\newcommand{\states}{S}

\newcommand{\post}{\mathit{succ}_t}

\newcommand{\bools}{\mathbb{B}}
\newcommand{\true}{\mathit{true}}
\newcommand{\false}{\mathit{false}}
\newcommand{\nats}{\mathbb{N}}

\newcommand{\SP}{\mathcal{SP}}

\newcommand{\game}{\mathcal{G}}
\newcommand{\gstates}{S}
\newcommand{\ginit}{s_0}

\newcommand{\spec}{\varphi}

\newcommand{\alphabet}{\Sigma}

\newcommand{\beq}{\begin{equation*} \begin{aligned}}
\newcommand{\eeq}{\end{aligned} \end{equation*}}

\DeclareCaptionLabelFormat{tablel}{#1Table #2}
\DeclareMathOperator*{\argmax}{arg\,max}

\IEEEoverridecommandlockouts                              

\title{\LARGE \bf Strategy Synthesis for Surveillance-Evasion Games\\ with Learning-Enabled Visibility Optimization}

\author{Suda Bharadwaj$^{1}$, Louis Ly$^{2}$, Bo Wu$^{1}$, Richard Tsai$^{2}$, and Ufuk Topcu$^{1}$  
\thanks{The University of Texas at Austin\\$^{1}$ {\tt\small \{sudab,bwu3,utopcu\}@utexas.edu}\\$^{2}$ {\tt\small \{louis,ytsai\}@oden.utexas.edu}}%
}

\begin{document}
\maketitle
\thispagestyle{empty}
\pagestyle{empty}

\begin{abstract}
This paper studies a two-player game with a quantitative surveillance requirement on an
adversarial target moving in a discrete state space and a secondary objective to
maximize short-term visibility of the environment. We impose the surveillance requirement as a temporal logic constraint. We then use a greedy approach to determine vantage points that optimize a notion of information gain, namely, the number of newly-seen states. 
By using a convolutional neural network trained on a class of environments, we can efficiently approximate the information gain at
each potential vantage point.  
Subsequent vantage points are chosen such that moving to that location will not jeopardize the surveillance requirement, regardless of any future action chosen by the target. 
Our method combines guarantees of correctness from formal methods with the scalability of machine learning to provide an efficient approach for surveillance-constrained visibility optimization.


\end{abstract}


\section{Introduction}

Over the last decade, the use of autonomous agents, such as unmanned aerial vehicles, for patrolling and surveillance in adversarial environments has increased tremendously \cite{gupta2016survey,kanistras2015survey}. In such settings, it is often necessary to not only perform routine patrolling of the area but also to maintain visibility of potentially hostile targets until an appropriate response can be formed. Thus, an autonomous agent has two objectives. Throughout the operation, it must maintain line-of-sight visibility of, or otherwise track, a moving target. Second, the agent may also need to maximize its visibility of the environment, prioritizing areas which it has not recently observed.
Simply, the agent needs to \emph{optimally patrol} an environment subject to the constraint of a surveillance requirement on a moving target. 

There is significant amount of existing work on designing strategies for autonomous patrolling  agents in known environments~\cite{oyekan2009towards,semsch2009autonomous,7165356,liu2017communication,liu2017distributed} and unknown environments~\cite{landa2008visibility,valente2014information,yamauchi1997frontier,ghosh2008online,gonzalez2002navigation, bircher2018receding,heng2015efficient,surmann2003autonomous}. 
However, in these cases, the agents are merely \emph{passive observers}. When a hostile target is detected, the agents simply inform a human operator (for example via a live camera feed) who will take over from then on. However, in practice, targets could be lost after detection and thus need to be actively tracked by the autonomous agent.  In this paper, we treat the patrol problem as finding and visiting a set  of  discrete locations that provides visibility of the whole map. Such an optimization problem can be seen as an instance of the \emph{art gallery problem}, which has been shown to be NP-hard \cite{urrutia2000art}. 


The surveillance of adversarial targets is naturally formulated as a two-player game.
There have been several variants of such games studied in~\cite{Basilico12, Chung2011}, including the case in which the environment is not known apriori~\cite{burger2009discovering,landa2011discovery}. These settings, however, only handle a simple surveillance requirement, namely to never lose sight of the target. For example, in~\cite{takei2014efficient} the authors formulate the problem as a two-player game in which both players choose their controls at initial time and proceed until the target is able to hide. If this requirement is relaxed, the agent may not always observe, or even know, the exact location of the target. In this case, surveillance is, by its very nature, captured by a partial-information two-player game. 

While there has been a lot of work on both surveillance and patrolling individually, there has been very little on the combination of the two problems. To bridge this gap, this paper proposes a method for maximizing the visibility of the environment for the patrol objective while \emph{actively maintaining knowledge} of the location of a hostile target for the surveillance objective. We provide a \emph{quantitative guarantee} on surveillance performance.

Over an infinite time horizon, the agent must be able to prevent its uncertainty of the target's location from exceeding a user-defined threshold. The approach taken in this paper is closely related to the framework established in~\cite{Bh2018} and~\cite{bharadwaj2018distributed} where a surveillance game is formulated as a GR(1) synthesis problem. However, no patrolling objective is considered in ~\cite{Bh2018} nor~\cite{bharadwaj2018distributed}. Another salient difference is that we do not want to synthesize a single strategy, but instead generate a \emph{winning region}. Within the winning region, the agent is guaranteed to not violate the surveillance specification. We use the winning region as a constraint for the patrol task to guarantee correctness of the surveillance requirement.

For maximizing visibility in large-scale environments, we leverage the efficiency of data-driven techniques to determine approximately-optimal  locations for the agent to move towards. We adopt the approach from~\cite{ly2019autonomous}, where a neural network is used to  approximate a gain function that quantifies map visibility. However, the approach in~\cite{ly2019autonomous} does not consider any surveillance requirements and focuses purely on visibility optimization. We constrain the generation of these locations at runtime to guarantee the correctness with respect to a surveillance specification. The concept of guaranteeing correctness on systems with different goals is studied in the field of runtime verification \cite{multiagentshield,bloem2015shield} where a winning set is computed offline and actions that take the system out of the set are overwritten at runtime. We employ this in principle as new locations are computed greedily at runtime, such that those new locations must stay in the winning region. 

Our contribution is summarized as follows:
\begin{itemize}
    \item[1)] We propose a \emph{scalable} algorithm to approximately-optimally explore an environment while guaranteeing a quantitative surveillance specification.
    \item[2)] We employ machine learning to generate locations that maximize visibility in large-scale discrete environments. Formal verification techniques are often not able to handle similar state space sizes~\cite{ray2010scalable}. We develop an abstraction method to reduce the state space of a discrete environment. We show that the abstraction is sound which guarantees that if a strategy is correct on an abstract environment, it is also correct on the underlying concrete environment. We are thus able to leverage the scalability of machine learning with the guarantees of formal methods. 
    \item[3)] We present a simulation on a case study and benchmark on various sizes of state spaces to demonstrate the scalability of the proposed approach.
\end{itemize}






The paper is structured as follows. Section II presents the problem informally in the context of a case study. Section III provides notations and definitions for the surveillance game structures. The patrol problem constrained with a surveillance specification is formally constructed in Section IV and the solution approach is given in Section V. We present a method to generate a sound abstraction of a large discretized state space in Section VI and provide numerical evaluations in Section VII. Finally, we conclude in Section VIII and discuss avenues of future research.

\section{Game structure}

\subsection{Environment}
\begin{figure}[h!]
\centering
\subfloat[Surveillance arena \label{fig:map}]{
\includegraphics[width=0.4\linewidth]{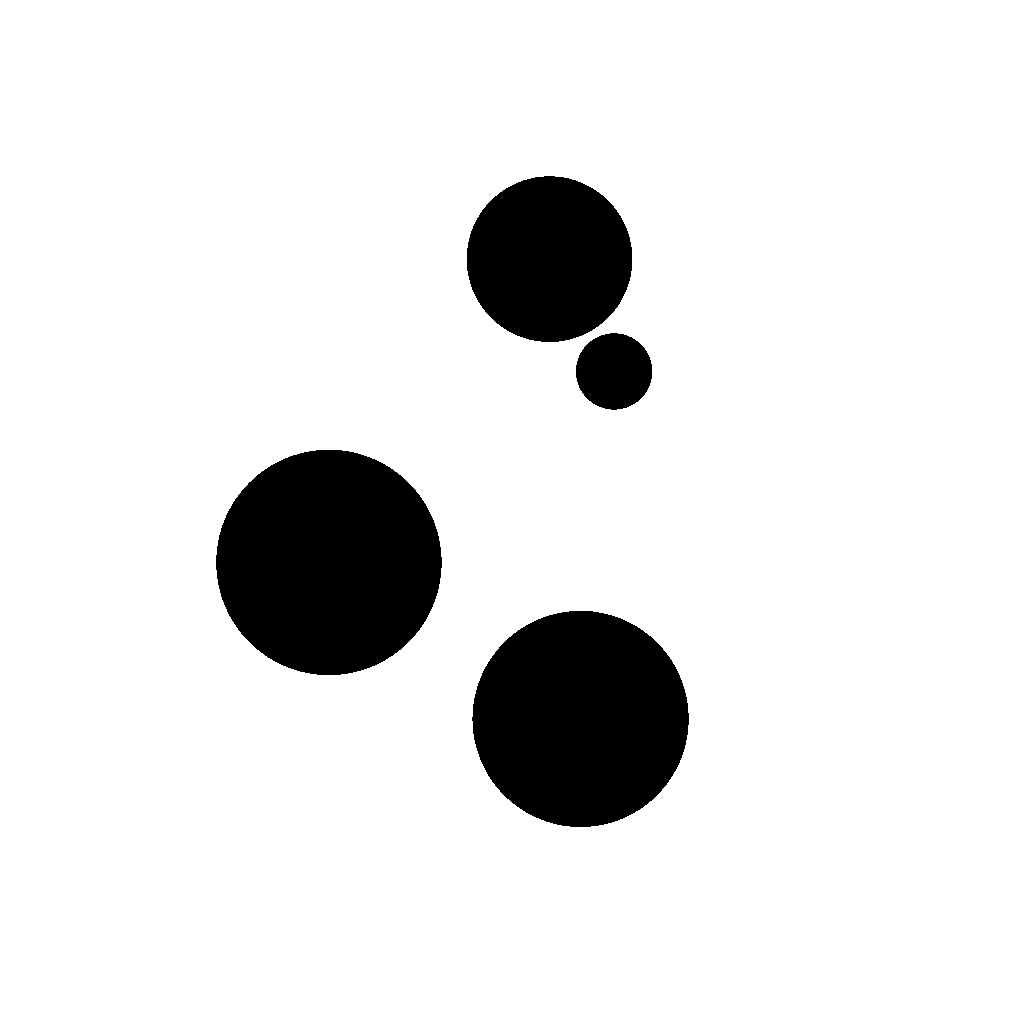}
}
\subfloat[Gridworld representation\label{fig:gridofmap}]{
\includegraphics[width=0.4\linewidth]{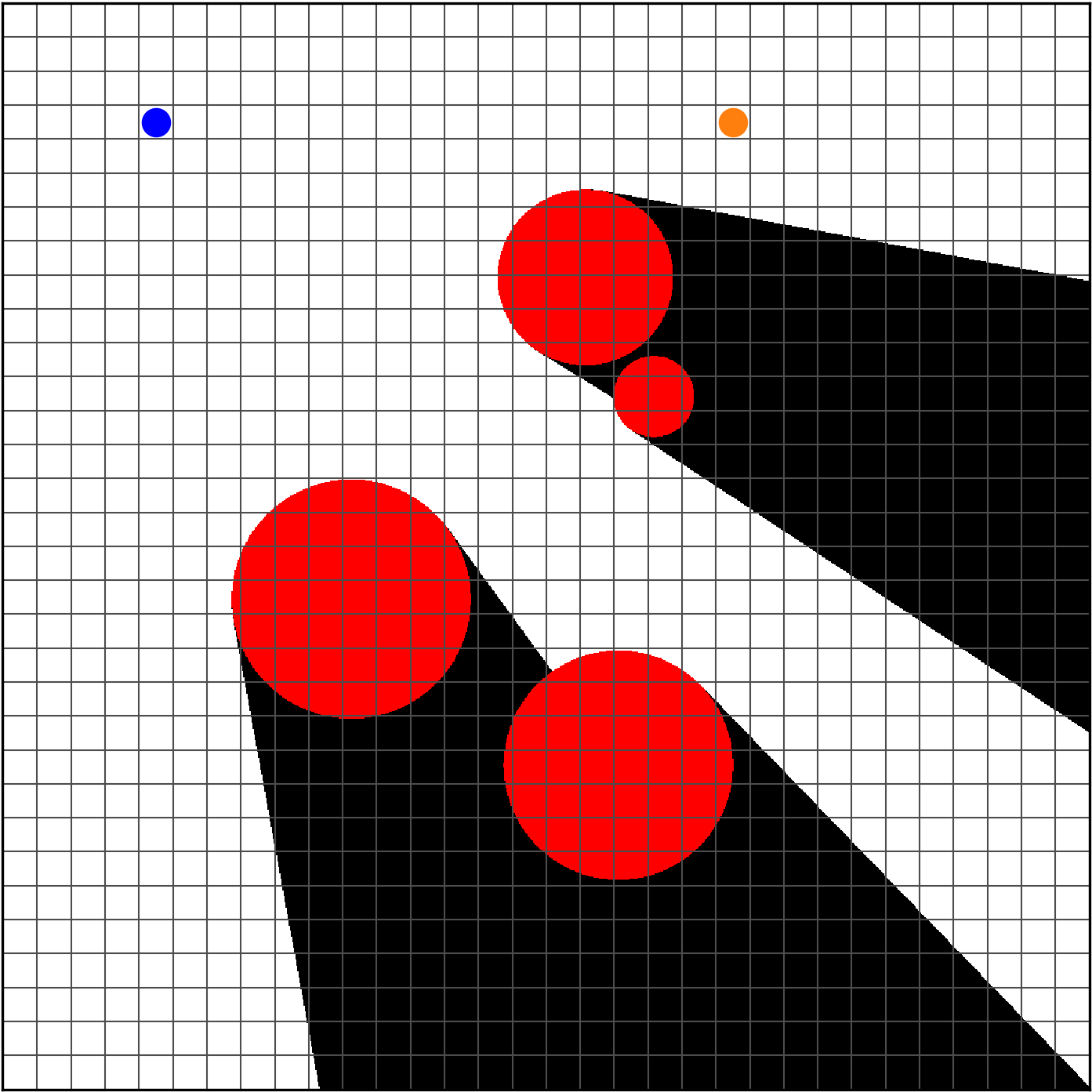}
}
\caption{Sample environment and corresponding gridworld representation. The blue circle corresponds to the controlled agent and the orange circle corresponds to the hostile target. Red cells are obstacles that obscure vision. Black cells in (b) correspond to occlusions.}\label{fig:casestudy}
\end{figure}

We represent the environment, an example of which is shown in Figure~\ref{fig:map}, as a set of states $\mathcal L$. We define $L \subseteq \mathcal{L}$ as the set of states representing free space and $L^c = \mathcal L\setminus L$ representing the obstacles. For example, in Figure~\ref{fig:gridofmap}, $L$ corresponds to all the white cells and $L^c$ corresponds to the obstacles in red.

\subsection{Surveillance games}\label{sec:surveillance-games}
We define a \emph{surveillance game} to be  a tuple $G = (\game,\vis)$ with $\game  = (\gstates,
\ginit, \delta)$, where:
\begin{itemize}
\item $\gstates = L_a \times L_t$ is the set of states, with $L_a$ the set of locations of the agent, and $L_t$ the locations of the target;
\item $s^\init = (l_a^\init,l_t^\init)$ is the initial state;
\item $\delta \subseteq \states \times \states$ is the transition relation describing the possible moves of the agent and the target; and
\item $\vis : \states \to \bools$ is a function that maps a state $(l_a,l_t)$ to $\true$ iff \emph{ position $l_t$ is in sight of $l_a$}.
\end{itemize}

The relation $\delta$ captures the next allowed move of both the target and the agent, where the target moves first and the agent moves second. For a state $(l_a,l_t)$ we define $\succs_t(l_a,l_t)$ as the set of possible successor states of the target. We extend $\succs_t$ to sets of states of the target by stipulating that the set $\post(l_a,L)$ consists of all possible successor states of the target for states in $\{l_a\} \times L$. Formally, let $\post(l_a, L) = \bigcup_{l_t \in L}\succs_t(l_a,l_t)$.

For a state $(l_a,l_t)$ and a successor location of the target $l_t'$, we denote with $\succs_a(l_a,l_t,l_t')$ the set of successor locations of the agent, given that the target moves to $l_t'$: 

$\succs_a(l_a,l_t,l_t') = \{l_a' \in L_a \mid  \delta(l_a,l_t) = (l_a',l_t') \}$.



\subsection{Belief-set game structures}

For surveillance games, when the agent cannot see the target, we need to reason about the \emph{belief} the agent has in the location of the target. 
 To this end, we employ a powerset construction which is commonly used to transform a partial-information game into a perfect-information one, by explicitly tracking the knowledge the agent has as a set of possible locations of the target.

Given a set $B$, we denote with $\mathcal{P}(B) = \{B' \mid B'\subseteq B\}$ the powerset (set of all subsets) of $B$.

For a surveillance game $G = (\game,\vis)$ we define the corresponding \emph{belief-set game structure} $G_\belief  = (\game_\belief,\vis)$ where $\game_\belief = (\states_\belief,s^\init_\belief,\delta_\belief)$ where:
\begin{itemize}
\item $\states_\belief = L_a \times \beliefs$ is the set of states, with $L_a$ the set of locations of the agent, and $\beliefs$ the set of \emph{belief sets} describing information about the location of the target;
\item $s^\init_\belief = (l_{a_{init}},B_{t_{init}})$ is the initial state;
\item $\delta_\belief: \states_\belief \times \states_\belief$ is the transition relation where $\delta_\belief(l_a, B_t) = (l_a', B_t')$ iff $l_a' \in  \succs_a(l_a,l_t,l_t')$ for some $l_t \in B_t$ and $l_t' \in B_t'$ and one of these holds:
\begin{itemize}
\item[(1)] $B_t' = \{l_t'\}$, $l_t' \in \post(l_a,B_t)$, $\vis(l_a,l_t') = \true$;
\item[(2)] $B_t' = \{l_t' \in \post(l_a,B_t)  \mid  \vis(l_a,l_t') = \false \}$.
\end{itemize}
\end{itemize}




A \emph{run} $\rho$ in the game $G_\belief$ is an infinite sequence $ \rho = (l_{a_0},B_{t_0}),(l_{a_1},B_{t_1})\ldots$ of states in $\states_\belief$, where $(l_{a_0},B_{t_0})= (l_{a_{init}},B_{t_{init}}) = s_\belief^\init$,  $\delta_\belief((l_{a_i},B_{t_i}),\sigma_i) = (l_{a_{i+1}},B_{t_{i+1}})$ for all $i$.  A \emph{strategy for the agent in $G_\belief$} is a function $f_a : S_\belief \times \beliefs \to S_\belief$ that maps the history of the play and the current belief of the target location to the agent's next action. 


\subsection{Temporal surveillance objectives}
We consider a set of \emph{surveillance predicates} $\SP = \{p_k \mid k \in \nats_{>0}\}$, where for $k \in \nats_{>0}$ we say that a state $(l_a,B_t)$ in the belief game structure satisfies $p_k$ (denoted $(l_a,B_t) \models p_k$) iff 
$|\{l_t \in B_t \mid \vis(l_a,l_t)  = \false \}| \leq k$. Informally, a state in the belief game satisfies $p_k$ if the size of the belief set does not exceed the threshold $k \in \nats_{>0}$.

Of focus in this paper will be linear temporal logic (LTL) surveillance formulas of the form $\LTLglobally p_k$, termed \emph{safety surveillance objective} which is satisfied if at each point in time the size of the belief set does not exceed $k$. 

LTL formulas of this type are interpreted over (infinite) runs. If a run $\rho$ satisfies an LTL formula $\varphi$, we write $\rho \models \varphi$. The formal definition of LTL semantics can be found in~\cite{BaierKatoen08}.




\section{Problem Formulation}

\subsection{Patrol problem}
Consider an environment as in Figure \ref{fig:casestudy} represented by a set $\mathcal L$, with a set $L \subseteq \mathcal L$ of free states. We construct a surveillance game structure $G = (\game,\vis)$ with $\game  = (\gstates,
\ginit, \alphabet, \delta)$. Recall that $S = L_a \times L_t$ is the joint state space of the target and agent. We assume without loss of generality that $L_a = L_t = L$, i.e., both target and agent share the same state space, which is natural in the setting where both the agent and target can move in the same discretized space. 





Given the visibility function $\vis$, let $\mathcal{V}_{\vis}(l) \subseteq L$ be the set of points in $L$ that are visible from $l \in L$, i.e., $$\mathcal{V}_{\vis}(l) := \{l_t\in L\mid \vis(l,l_t) = true\}.$$

We first present the patrol problem studied in \cite{ly2019autonomous}. 
\begin{problem}
Find the smallest set $O \subseteq L$ of \emph{vantage points}  such that each point in $L$ is visible from at least one vantage point $l \in O$, i.e.,
\beq 
\min_O \ & \ |O| \qquad
\text{s.t. }& \bigcup_{l\in O} \mathcal{V}_{\vis}(l) = L.\\[1em]
\eeq 
\end{problem}

This problem was solved in \cite{ly2019autonomous}, by using a greedy approach to determine vantage points sequentially; i.e., a new vantage point is determined based on the visibility of the map obtained from the previous vantage points. Each new point is chosen to maximize information gain. In this case, since we deal with discrete state spaces, the information gain is quantified as the number of previously-unseen states that become visible after moving to a potential vantage point $l_i$. We define this \emph{gain function} as:
\begin{align}\label{eqn:gain}
     g^K(l_i;\Omega_{i-K}^{i}) = |\mathcal{V}_{\vis}(l_i) \cup \Omega_{i-K}^{i}| - |\Omega_{i-K}^{i}|,
\end{align}
where $\Omega_{i-K}^{i}$ represents the set of visible points for the last $K$ observations:
$$\Omega_{i-K}^{i} = \bigcup_{j=i-K}^{i} \mathcal{V}_{\vis}(l_j).$$
We enforce that $i - K = 0$ when $i \leq K$.
When $K = \infty$, the agent remembers everything it sees and patrolling is complete when all the states in the map have been seen at least once. When $K$ is finite, the agent has to try to observe states it has not seen in the last $K$ steps. In the next section, we discuss the patrol problem with finite $K$.

\begin{figure}[h!]
\centering
\subfloat[Obstacles and visibility \label{fig:gainexample0}]{
\includegraphics[height=0.45\linewidth]{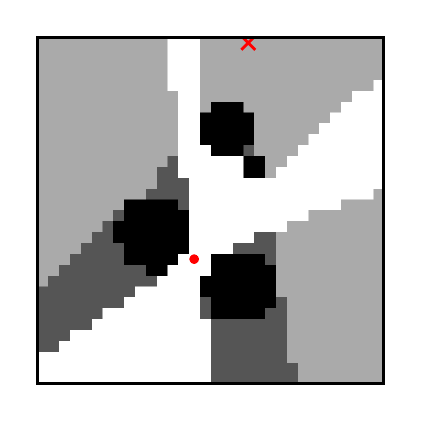}
}
\subfloat[Gain function\label{fig:gainexample1}]{
\includegraphics[height=0.45\linewidth]{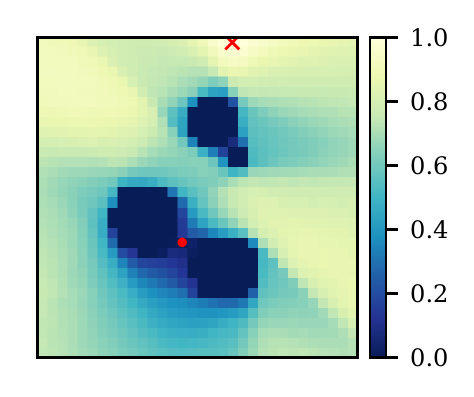}
}
\caption{(a) Black regions are obstacles. White regions are visible from the red dot.  Light gray regions are visible from the cross, but not from the dot. The area of light gray region corresponds to the gain at the cross. (b) The gain function (normalized for visualization), given that the dot has been visited. The cross marks the location of maximum information gain.}\label{fig:gainexample}
\end{figure}

\begin{example}
Figure~\ref{fig:gainexample} shows a visualization of the gain function for the environment shown in Figure~\ref{fig:map}.  
\end{example}

Unfortunately, computing $g^K(l;\Omega_{i-K}^{i})$ is costly. For each $l$, computing the gain function requires $\mathcal O(|L|)$ operations. Thus computing the gain for all $l \in L$ has complexity $\mathcal O(|L|^2)$ \cite{ly2019autonomous}. The computation becomes prohibitively expensive, especially for large state spaces. It is worth noting that this computation cannot be performed offline as it will depend on the choices of the target at run time. 




\subsection{Constrained patrol problem}
We first provide some additional notation. We note that to help readability and avoid nested subscripts, we slightly abuse notation by dropping the subscripts $a,t$ in the state $s_i = (l_{a_i},B_{t_i})$. The state in the belief-set game at time $i$ is referred to in this section as $(l_{i},B_{i})$.
Given a belief-set game $G_\belief  = (\game_\belief,\vis)$, let $\rho = s_0,s_1,\dots = (l_0,B_0),(l_1,B_1),\dots$ be an infinite run in the belief game $\game_{\belief}$. We define $\rho^i_j = ( (l_j,B_j),\dots (l_i,B_i))$ with $j<i$ as the finite \emph{word} in the infinite run from $j$ to $i$. We can thus write the infinite run as $\rho = \rho^{j-1}_{0} \cdot \rho^i_j \cdot \rho^\omega_{i+1}$ and denote $\rho^{j-1}_{0} = (l_0,B_0),\dots (l_{j-1},B_{j-1}) $ as a \emph{finite prefix} and $\rho^\omega_{i+1} = (l_{i+1},B_{i+1}),(l_{i+2},B_{i+2}),\dots$ as an \emph{infinite suffix}.
We are interested in constraining the patrol problem with the addition of a surveillance requirement. However, recall that the temporal surveillance objectives are only satisfied over infinite runs. This setting differs from the approach in~\cite{ly2019autonomous} in that it now involves an infinite time horizon. 
We thus want to maximize the short-term visibility in a time interval of length $K$ in order to incentivize the agent to gain visibility of states it has not seen from the last $K$ vantage points. 

We construct a \emph{sequentially optimal} vantage point generation problem. Simply, given the current state of the belief game $(l_{i},B_{i})$ and an integer $K > 0$, we want to choose the next vantage point that maximizes the visibility of states that have not been seen by the last $K$ vantage points and does not violate the surveillance safety specification $\spec$.

We now present the main problem of this paper. 
\begin{problem}\label{prob:main}
Given a belief-set surveillance game $G_\belief$, at time step $i-1$, the run in the game is $\rho_0^{i-1}$. Additionally, given $K \in \mathbb{N}$, and a surveillance requirement $\varphi$, solve
\beq 
\argmax_{l_i} \ &  g^K(l_i;\Omega_{i-K}^{i}) \qquad
\text{s.t. }& \rho \models \spec
\eeq 
where $\rho = \rho_{0}^{i-1}\cdot (l_i, B_i)\cdot  \rho_{i+i}^{\omega}$ for all $B_i \in \succs_t(l_{i-1},{B_{i-1}})$. 
\end{problem}

Informally the agent needs to choose the next vantage point that guarantees the infinite suffix will still be correct with respect to the surveillance specification regardless of the target's action choices. 

In the next section we present how to translate the constraint on the infinite suffix as a feasible set for the optimization problem. The resulting optimization problem  needs to be solved online for every time step. As discussed previously, evaluating the exact gain function for all feasible points is a computationally intensive task. We thus show how to approximate the gain function using a convolutional neural network which allows for the generation of the next vantage point in real time even for large maps. 

\section{Solution approach}

\subsection{Surveillance winning region}
Consider a pair $(G,\varphi)$, where $G$ is a surveillance game structure and $\varphi$ is a surveillance objective. Given the relevant surveillance predicates and specification over the belief-set game structure $G_{\belief}$, we obtain a \emph{safety game}. 

A safety game defines a set $F\subseteq L\times \mathcal{P}(L)$ of safe states. A run $\rho \models \spec$ iff $s_i \in F$ for all $i \geq 0$, i.e., if only safe states are visited in the run. Otherwise, $\rho \nvDash \spec$. The \emph{winning region} $\mathcal W \subseteq F$ is the set of states from which a winning strategy exists.  Informally a state $(l_a,B_t)$ in the belief game is said to be in the winning region $\mathcal W$ iff, for all future steps the target can make from the current belief state $B_t$, there exists a strategy from $l_a$ that guarantees that only states in $\mathcal{W}$ will be visited. We can use standard algorithms for safety games (e.g.~\cite{Mazala01}), to compute the winning region.

We note that to avoid the state space blow-up arising from the subset construction in the partial information game, we solve the surveillance games using the \emph{belief-set abstraction} techniques detailed in \cite{Bh2018}. 

\subsection{Feasible set construction}
  Synthesizing a strategy for a surveillance specification is inherently \emph{reactive}, the agent observes the target's action and reacts appropriately in order to satisfy its specification. However, planning only the immediate action, as is done in \cite{Bh2018}, will not result in a large gain in visibility coverage.


Informally, we want to be able to move to the best vantage point possible in a region that guarantees that no matter what the target does in the future, there will always exist a way for the agent to satisfy its surveillance requirement. 





More concretely, consider the belief-set game $G_\belief$ along with a surveillance specification $\varphi$. Assume we have solved  the resulting safety game and computed the winning region $\mathcal W \subseteq L\times \mathcal{P}(L)$.
Recall that the state of the belief-set game structure is the joint position and belief $s = (l_a,B_t)$. We extend the definition of $\succs_a(l_a,l_t,l_t')$ to sets by stipulating that the set $\succs_a(l_a,B_t) = \bigcup_{l_t' \in \succs_t(l_a,B_t)} \succs_a(l_a,B_t,l_t')$. Additionally, we extend the definition of $\succs_t(l_a,B_t)$ to more than the immediate successor by recursively defining $\succs_t^k(l_a,B_t) = \bigcup_{B_t' \in \succs_t^{k-1}(l_a,B_t)} \succs_t(l_a,B_t)$.

We now define the \emph{reachable} successor set $R(l_a,B_t)$ as 
\begin{multline}\label{eqn:R}
R(l_a,B_t) = \{l_a' \in L \mid l_a \in \succs_a(l_a,B_t), \\ \forall B_t' \in \succs_t(l_a,B_t), (l_a',B_t') \in \mathcal{W} \}.  
\end{multline}

Informally, a successor state $l_a'$ is a reachable successor of $(l_a,B_t)$ if, \emph{for all possible successor belief states} $B_t'$, we have that $(l_a',B_t')$ does not violate the surveillance specification. A state being in the set $R(l_a,B_t)$ a strict condition -- the same successor state $l_a'$ has to be correct for all possible $B_t'$ choices the target can make. 

We can then recursively define reachable successor set over a finite horizon $k$, by applying Equation~\ref{eqn:R} $k$ times:
\begin{multline*}
R^k(l_a,B_t) = \{l_a' \in \bigcup_{\l_a \in R^{k-1}}\succs_a(l_a,B_t) \mid \\ \forall B_t' \in \succs_t^k(l_a,B_t), (l_a',B_t') \in \mathcal{W} \}.    
\end{multline*}
Intuitively, as $k$ gets larger, finding a point $l \in L$ that guarantees staying in the winning region $\mathcal W$ gets harder. We define $k^\ast$, the largest $k$ such that $R^{k^\ast+1} = R^{k}$ and refer to the set as the \emph{maximum reachability set} $R^\ast(l_a,B_t)$.

\paragraph*{Remark}Computing the set $R^\ast$ is similar to standard fixed point set computations detailed in~\cite{gradel2002automata} for computing the winning region $\mathcal W$ in temporal logic synthesis problems. The main difference is that these algorithms focus on computing a set where \emph{for all} actions by a player \emph{there exists} a strategy for the other player to win. We are interested in reversing the order of the quantifiers which results in a more restrictive condition. One can think of $k^\ast$ as the maximum number of actions the agent can make without ''reacting" to the target's actions and still guaranteeing correctness with regards to the specification. 



The reachable set computation can be performed online in $\mathcal O(|L \times \mathcal{P}(L)|^2)$. However, if required, the entire computation can be performed offline with complexity $\mathcal O(|L \times \mathcal{P}(L)|^4)$, and the results stored for all possible states -- there will be a set $R^\ast$ stored for up to $|L \times \mathcal{P}(L)|$ states. 

We now present an \emph{equivalent} version of Problem \ref{prob:main} with the constraint on the infinite run translated to a feasible set. Given that the state of the game at time $i$ is $(l_i,B_i)$, the next vantage point is the solution to the following problem:

\begin{problem}\label{prob:maingreedy}
Given a belief-set surveillance game $G_\belief$ at time $i-1$,  $K \in \mathbb{N}$, and a surveillance requirement $\varphi$, solve

\beq 
\argmax_{{l \in R^{\ast}(l_{i-1},B_{i-1})}} \ &  g^K(l_i;\Omega_{i-K}^{i})\\[1em]
\eeq 
\end{problem}
where $(l_{i-1},B_{i-1})$ is the state in the game at $t = i-1$.

\subsection{Approximating the gain function}
When the state space is large, computing the gain function is computationally expensive. In such cases, we approximate $g^K(l_i;\Omega_{i-K}^{i})$ using a convolutional neural network which enables efficient computation of the gain function at run time.

Without loss of generality, we discuss the training process for $K=\infty$.
The approximated gain function $g_\theta(l_i;\Omega_{0}^{i},B_i)$ takes as input the cumulatively visible set $\Omega_{0}^{i}$, and the associated shadow boundaries $\mathcal B_i= \partial \Omega_{0}^{i} \setminus L^C$. Here, $\partial \Omega_{0}^{i}$ is boundary of the set of states visible from at least one of the previous vantage points. Intuitively, the shadow boundaries (or frontiers) $\mathcal{B}_i$ are the boundaries between free space and occlusion.
In order to train the neural network, we sample different environments $\mathcal L$ cropped from the INRIA Aerial Image Labeling Dataset~\cite{maggiori2017dataset}. For each $\mathcal{L}$, an initial position $l_0 \in L\subseteq \mathcal L$ for the agent is randomly sampled. We train on 100k corresponding data pairs $\{(\Omega_{0}^{i}, \mathcal B_i), g(l_i; \Omega_{0}^{i}) \}$.

Each subsequent $l_i$ after the initial position is chosen as
$$  l_{i} = \arg\max_l g(l; \Omega_{0}^{i-1}).$$

The parameters $\theta$ of the neural network are updated to minimize the cross entropy loss $E\Big(g(l_i; \Omega_{0}^i),g_\theta(l_i;\Omega_{0}^i, \mathcal{B}_i) \Big)$ between the normalized exact gain function $g$ and the normalized prediction $g_\theta$, where $E$ is defined as 
$$ E(p,q) :=\int p(x) \log q(x) + (1-p(x)) \log (1-q(x)) \ dx.$$

The architecture of $g_\theta$ is based on U-Net \cite{ronneberger2015u}, a standard CNN for dense inference problems which aggregates information across various scales using skip connections. The network has nineteen 3x3 convolution layers, each followed by batch norm and leaky ReLU. Six of the convolutions use stride 2 for downsampling, while six are preceded by bilinear upsampling. For more details we refer the reader to~\cite{ly2019autonomous}.

At run time, we approximate the gain function using $g_\theta$, which is measure of how useful it is to place a new vantage point at $l$. Since not all $l$ are reachable due to the surveillance constraint, we restrict $g_\theta$ to be zero for $l$ outside $R^\ast$. The next vantage point is chosen as the point within the $R^\ast$ that maximizes $g_\theta$ and this process is repeated for all time.

\section{Surveillance game abstraction}

In order to scale to large state spaces, we present a conservative abstraction of the surveillance game $G$ with visibility function $\vis$. We first define an abstraction function.

\paragraph*{\textbf{Abstraction function}}
Given an environment with state space $L$, an \emph{abstraction function} $\psi\colon L\rightarrow 2^L$ yields a partition ${L}_\psi={l_1},\ldots,{l_n}$ of the state space $L$ with $\bigcup_{i=1}^n {l_i}=L$ and $\forall \;  i\neq j$ we have ${l_i}\cap{l_j}=\emptyset$.

\begin{theorem}
Given \emph{concrete} surveillance game $G = (\game,\vis)$, we say the abstract surveillance game $G_\psi= (\game_\psi,\vis_\psi)$ with $\game_\psi  = (\gstates_\psi,{\ginit}_\psi, \alphabet, \delta_\psi)$, and $ s_\psi = (l^a_\psi, l^t_\psi)$ with ${l^a_\psi,l^t_\psi}\in L_\psi$, generated by abstraction function $\psi$ is \emph{sound} if the following holds 

\begin{enumerate}
    \item $\delta_\psi(s_\psi,\sigma) = s_\psi'$ with $s_\psi' = (l^{a'}_\psi,l^{t'}_\psi)$ iff
    \begin{itemize}
        \item \emph{For all} $l^a \in l^a_\psi$ in the concrete game, it must be the case that $\delta((l_a,l_t),\sigma) = (l_a',l_t')$ with either $l_a' \in l^{a'}_\psi$ or $l_a' \in l^{a}_\psi$. Or equivalently, we require there does not exist $l^a \in l^a_\psi$ such that $l_a' \notin l^{a'}_\psi$ and $l_a' \notin l^{a}_\psi$.
        \item \emph{There exists} an $l^t \in l^t_\psi$ in the concrete game, such that $\delta((l_a,l_t),\sigma) = (l_a',l_t')$ with $l_t' \in l^{t'}_\psi$.
    \end{itemize}
    \item $\vis_\psi((l^a_\psi, l^t_\psi)) = \top$ iff \emph{for all} $(l_a,l_t) \in (l^a_\psi, l^t_\psi)$ we have $\vis(l_a,l_t) = \top$.
\end{enumerate}
\end{theorem}

Simply, the abstract version of the surveillance game gives more power to the target and removes power from the agent by disallowing transitions between abstract states for the agent if it is not possible \emph{for all} underlying concrete states. Conversely transitions between abstract states are allowed for the target if \emph{there exists} a corresponding transition in any of the underlying concrete states.

\begin{example}
\begin{figure*}[ht!]
    
	\begin{minipage}{5.0cm}
		\subfloat[$t=0$ \label{fig:case1t0}]{
			\includegraphics[width=0.54\linewidth]{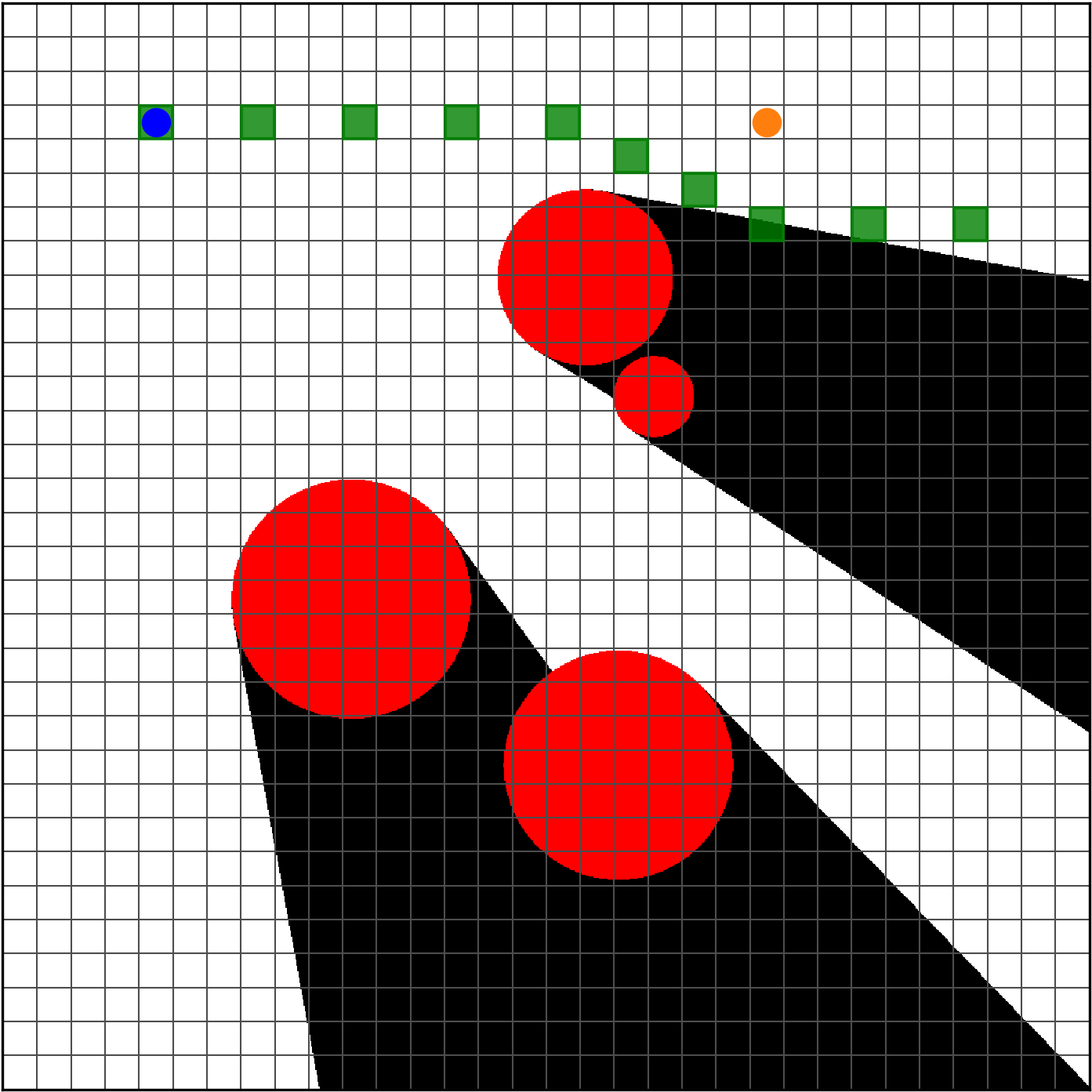}
		}
			\subfloat[$t=10$ \label{fig:case1t10}]{
			\includegraphics[width=0.54\linewidth]{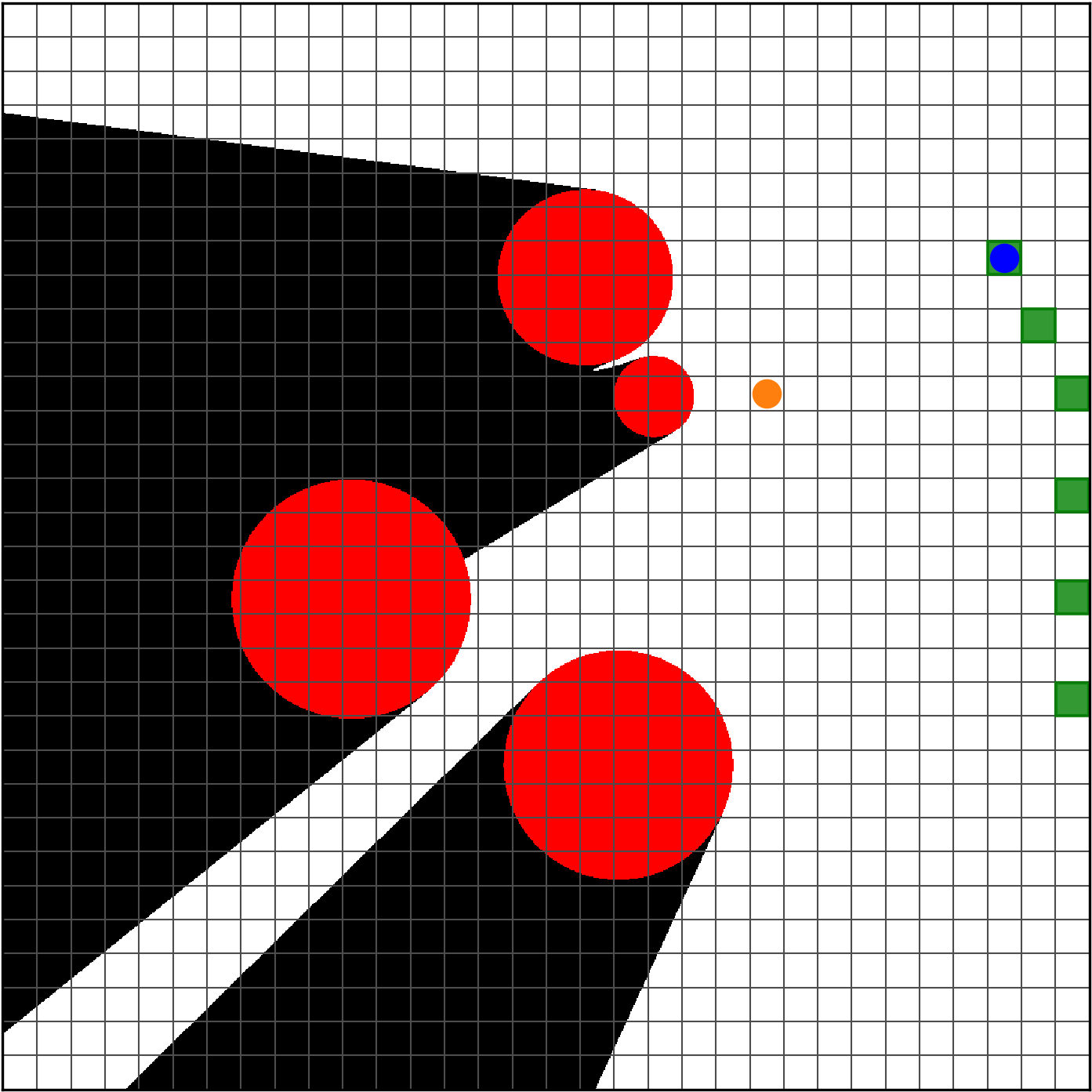}
		}
			\subfloat[$t=16$ \label{fig:case1t16}]{
			\includegraphics[width=0.54\linewidth]{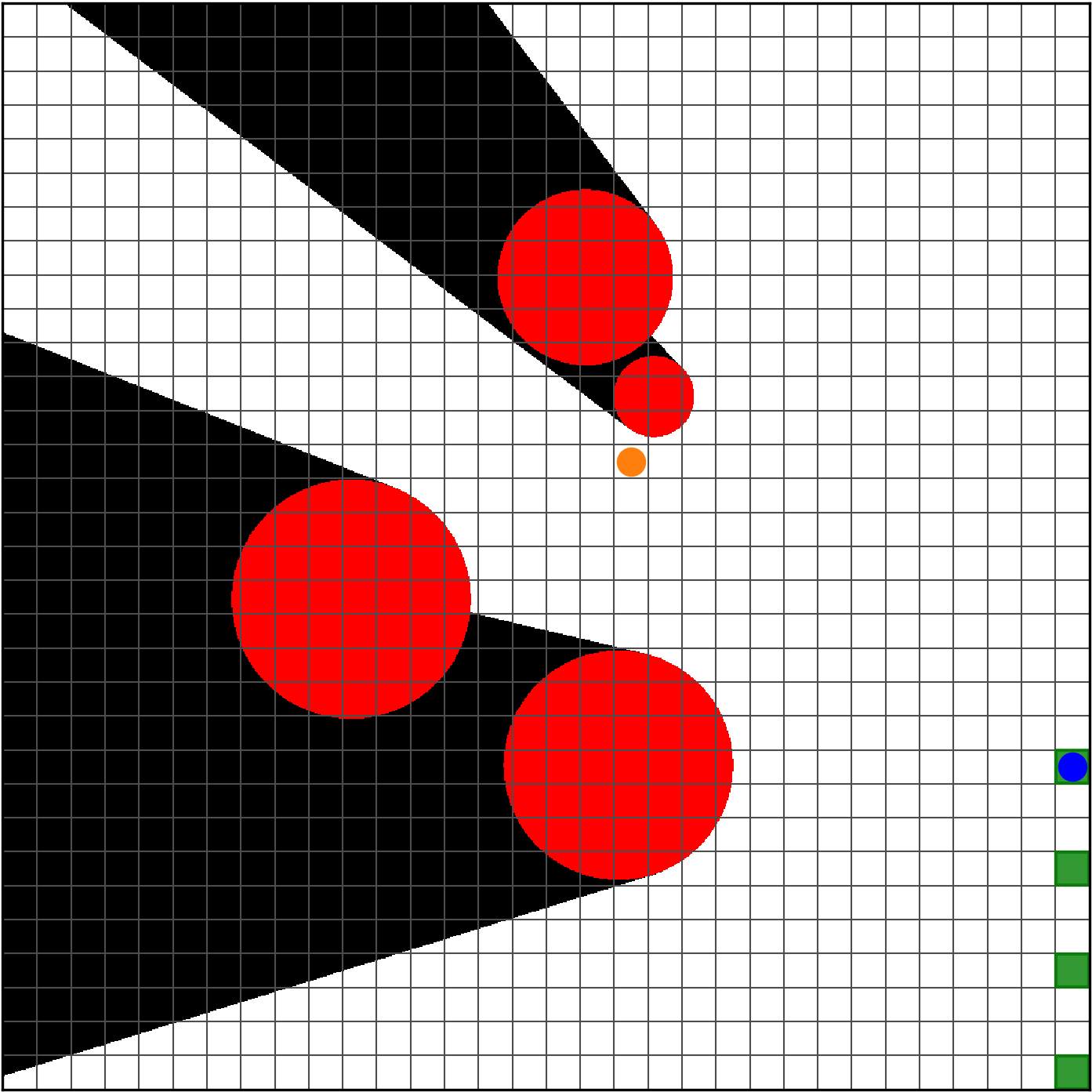}
		}
			\subfloat[$t=20$ \label{fig:case1t20}]{
			\includegraphics[width=0.54\linewidth]{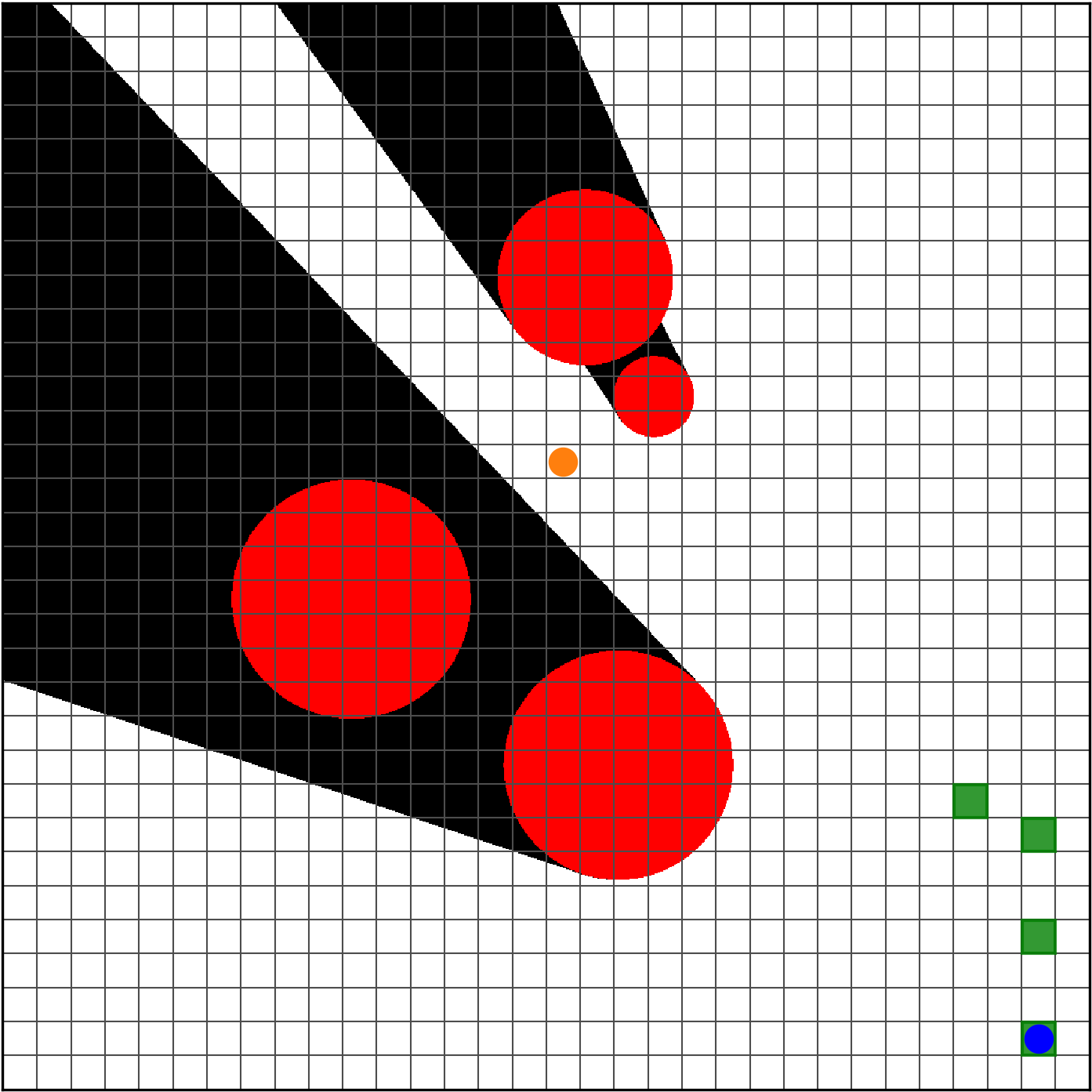}
		}
			\subfloat[$t=24$ \label{fig:case1t24}]{
			\includegraphics[width=0.54\linewidth]{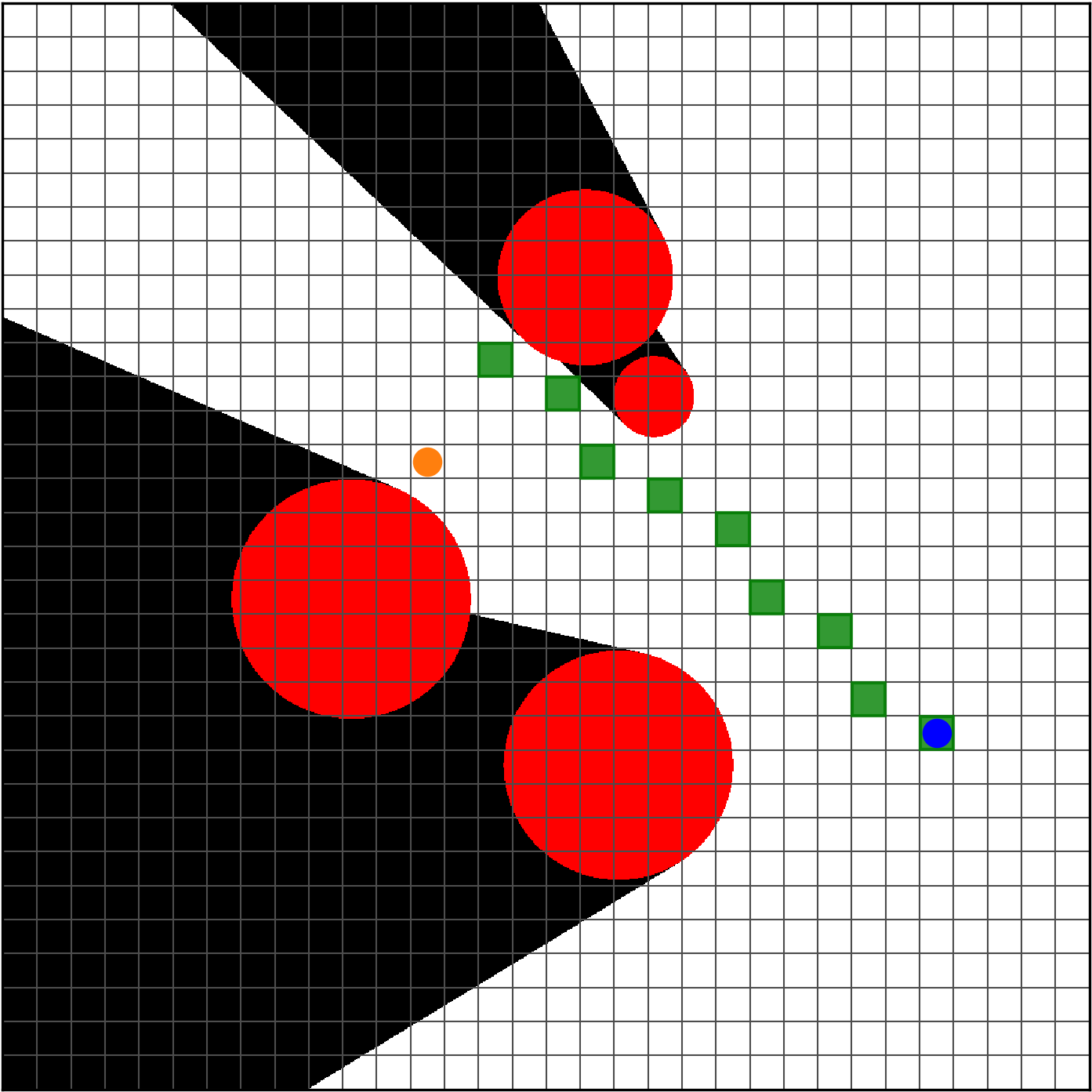}
		}
			\subfloat[$t=33$ \label{fig:case1t33}]{
			\includegraphics[width=0.54\linewidth]{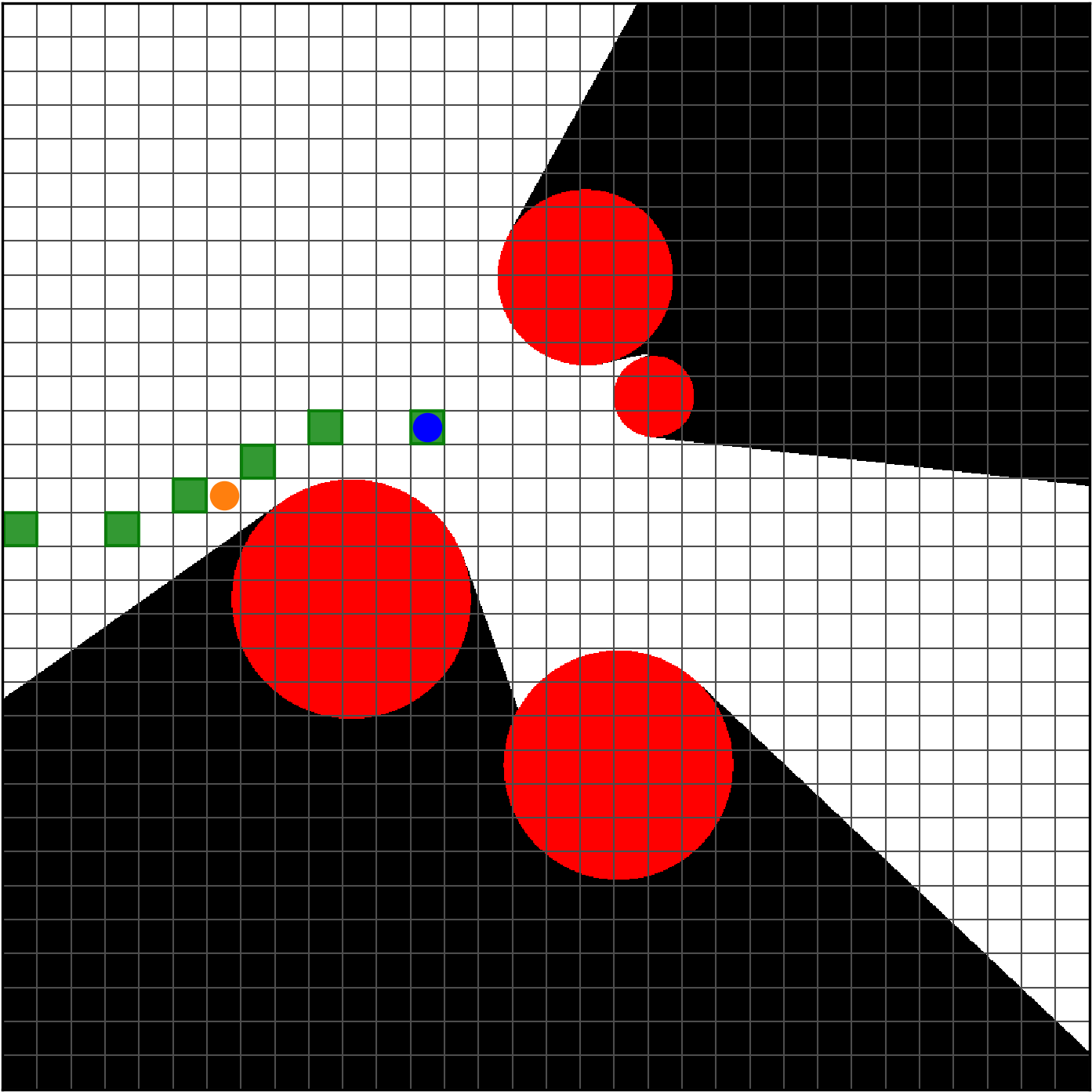}
		}

	\end{minipage}

	\caption{Simulation of the case study presented in Section II. The green cells correspond to the path to the optimal vantage point which is the last green square on the path. Snapshots of when a new vantage point is recomputed is shown here.}
	\label{fig:gridresults}
\end{figure*}

Figure~\ref{fig:casestudy} shows an example of abstracting a concrete map with a large state space to a coarser representation. Each grid state in Figure~\ref{fig:gridofmap} represents a collection of pixels from Figure~\ref{fig:map}.
\end{example}

This construction of the abstraction guarantees that a strategy that is correct on the abstract game with respect to safety surveillance objective $\spec$ corresponds to a winning strategy on the concrete game. However, since the abstraction is conservative, not finding a winning strategy on the abstraction does not mean one does not exist in the concrete game. In the experiments that follow in the next section, we use these abstractions to reduce the size of the state space for the winning region computation in the safety game.

\paragraph*{Remark} The methodology detailed in this paper introduces several sources of conservatism  -- state abstraction shown in this section and the belief abstraction in~\cite{Bh2018} used in generating strategies. These allow for more scalable synthesis on large problems. However, due to the added conservatism, there is no guarantee that there exists a winning strategy in the abstract model, even if one does exist in the concrete models. Techniques such as \emph{counterexample guided abstraction refinement} (CEGAR) can be used to generate new abstractions if no winning strategy can be found.

\section{Numerical Experiments}

\subsection{Effectiveness of patrol}
In order to illustrate the efficacy of including the vision-maximization for the patrol problem, we compare an agent employing the surveillance constrained patrol method with a window of $K = 5$ to an agent following a pure safety surveillance strategy. We compare the \emph{receding gain} --- the ratio of newly seen states (that have not been seen in the last $K = 5$ steps), to the total number of traversable states in the map.

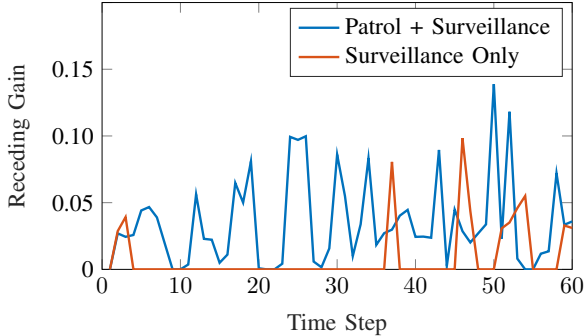
\begin{figure}[ht]
\centering
%
%
\definecolor{mycolor1}{rgb}{0.00000,0.44700,0.74100}%
\definecolor{mycolor2}{rgb}{0.85000,0.32500,0.09800}%

\begin{tikzpicture}[scale=0.9]

\pgfplotsset{scaled y ticks=false}
\begin{axis}[%
width=2.734in,
height=1.553in,
at={(1.297in,0.696in)},
scale only axis,
xmin=0,
xmax=60,
xlabel style={font=\color{white!15!black}},
xlabel={Time Step},
ymin=0,
ymax=0.2,
yticklabels={0.0, 0 , 0.05, 0.10, 0.15},
ylabel style={font=\color{white!15!black}},
ylabel={Receding Gain},
axis background/.style={fill=white},
legend style={legend cell align=left, align=left, draw=white!15!black}
]
\addplot [color=mycolor1, line width=1.0pt]
  table[row sep=crcr]{%
1	0\\
2	0.0269763635671602\\
3	0.0243860405908897\\
4	0.0257657840537603\\
5	0.0441412180650182\\
6	0.0467082809982365\\
7	0.0389531711896878\\
8	0.0194031050042925\\
9	0\\
10	0\\
11	0.00365076948451515\\
12	0.0562226958812807\\
13	0.0228836532646528\\
14	0.0222260284110853\\
15	0.0049353582258085\\
16	0.0110781241409643\\
17	0.0647644180548683\\
18	0.050074643592704\\
19	0.080925918877428\\
20	0.000816216056196264\\
21	0\\
22	0\\
23	0.00424178603297851\\
24	0.0993235556570539\\
25	0.0970535869100936\\
26	0.0997580955522569\\
27	0.00586470267321331\\
28	0.00156793835665681\\
29	0.0156973572361994\\
30	0.0861086793793375\\
31	0.0549455080627767\\
32	0.0100197921820881\\
33	0.0337704106877783\\
34	0.0832868132472289\\
35	0.0183775485606262\\
36	0.0271032365292632\\
37	0.0297813133043217\\
38	0.040248332677823\\
39	0.0445926743551682\\
40	0.0243680669212584\\
41	0.0245541472656762\\
42	0.02370515569427\\
43	0.0895268484333304\\
44	0.00265693128137463\\
45	0.0448326757084798\\
46	0.0286161966023421\\
47	0.0201632855022266\\
48	0.0269795353912128\\
49	0.0337249478763581\\
50	0.138853998824311\\
51	0.0231204827939118\\
52	0.118162075979988\\
53	0.00803105850112283\\
54	0\\
55	0\\
56	0.0117494935654263\\
57	0.0135415741551318\\
58	0.0720247233112152\\
59	0.0334426555356788\\
60	0.0359314801422669\\
};
\addlegendentry{Patrol + Surveillance}

\addplot [color=mycolor2, line width=1.0pt]
  table[row sep=crcr]{%
1	0\\
2	0.0286976067530249\\
3	0.0391900007189468\\
4	0\\
5	0\\
6	0\\
7	0\\
8	0\\
9	0\\
10	0\\
11	0\\
12	0\\
13	0\\
14	0\\
15	0\\
16	0\\
17	0\\
18	0\\
19	0\\
20	0\\
21	0\\
22	0\\
23	0\\
24	0\\
25	0\\
26	0\\
27	0\\
28	0\\
29	0\\
30	0\\
31	0\\
32	0\\
33	0\\
34	0\\
35	0\\
36	0\\
37	0.0805727891329079\\
38	0\\
39	0\\
40	0\\
41	0\\
42	0\\
43	0\\
44	0\\
45	0\\
46	0.0983286601792292\\
47	0.0436464135128163\\
48	0\\
49	0\\
50	0\\
51	0.0307286314213578\\
52	0.0351353523050703\\
53	0.0459058095129347\\
54	0.0550174027413018\\
55	0\\
56	0\\
57	0\\
58	0\\
59	0.0328653835581099\\
60	0.0310426420025628\\
};
\addlegendentry{Surveillance Only}

\end{axis}
\end{tikzpicture}%
\caption{Comparison of the effects of visibility optimization. Adding the patrol objective drastically increases the agent's visibility of the environment.}\label{fig:gain_compare}
\end{figure}

Figure \ref{fig:gain_compare} shows the results of the comparison. We note that when no patrol optimization is enforced, the agent tends to stay still as it is sufficient to maintain surveillance of the target. This causes no new states to be seen and results in the value of the receding gain to be zero for much of the time. Meanwhile, when visibility-maximization is taken into account, the agent actively moves to the best vantage point it can safely reach. The sum of gain over the time period is 2.07 when patrol optimization is performed vs 0.52 for the pure surveillance case. This corresponds to average gain of 3.4\% per time step vs 0.8\%. 

Finally, we remark that when the agent in the pure surveillance case stays still for longer than $K$ steps, all states not in its current vision become unseen states. Hence, when it finally moves, the gain becomes high, causing the spikes.

\subsection{Motivating case study}
Here we present the results of solving the case study presented in Section II with surveillance requirement $\square p\leq 1$ -- which corresponds to not allowing the agent to lose sight of the target. We assume the agent can move 4 cells for every 1 cell the target can move. The target is controlled by a human moving it with the arrow keys on a keyboard. 

We have implemented the simulation in Python, using the slugs reactive synthesis tool~\cite{EhlersR16} to solve the safety game and find the winning region $\mathcal{W}$. The experiments were performed on an Intel i5-5300U 2.30 GHz CPU with 8 GB of RAM. We solve for the winning surveillance region on the $32\times 32$ abstraction shown in Figure~\ref{fig:gridofmap}. New vantage points and the paths to those points were computed at each snapshot shown in Figure \ref{fig:gridresults} in green, and took approximately $\approx 1$ second for each new computation. Once the agent reaches the vantage point, the next one is computed and this process is repeated. We see that the agent can leverage its superior speed to move away from the target in order to see new areas, but it is still able to keep sight of the target at all times. 
A video of the simulation can be seen at \url{https://bit.ly/2F2Vnjb}.

\subsection{Computation benchmarks}
Here we compare the process on various sized maps in order to demonstrate the scalability of using a CNN to compute the gain function. See Table \ref{fig:runtime} for a comparison of computation time for the gain function. The CNN computes the gain function within 1 second for maps up to $1024\times1024$. 
Meanwhile, the exact gain function computation already takes more than 1 second for $128\times128$ maps and more than 2 hours for $1024\times1024$ maps. The scalability is crucial as the vantage point computation is conducted online on the concrete environment, compared to the surveillance strategy synthesis which is performed offline on an abstraction.

We measure the effectiveness of the gain function prediction by the decay of the residual, i.e. the number of steps required to reduce the occlusions to a certain threshold. The exact computation and CNN approximation generated sequences of vantage points that converged to the same level of coverage in a comparable number of steps. For a more detailed discussion on the evaluation of the gain function approximation, we refer the reader to~\cite{ly2019autonomous}.


\begin{table}[]
\caption{Comparison of gain function computation time (secs) for various map sizes. Using the exact gain function quickly becomes prohibitively expensive. }
\label{fig:runtime}

\centering{
\begin{tabular}{l|lllll}
               & \multicolumn{5}{c}{\textbf{Grid size}}                                                                                         \\ 
               & \textbf{$64^2$} & \textbf{$128^2$} & \textbf{$256^2$} & \textbf{$512^2$} & \textbf{$1024^2$} \\ \hline
\textbf{Exact} & 0.499                 & 2.565                   & 36.234                  & 597.797                 & 8105.750                  \\
\textbf{CNN}   & 0.058                 & 0.071                   & 0.126                   & 0.334                   & 1.194                    
\end{tabular}
}
\end{table}

\subsection{Discussion}
We are able to generate approximately optimal vantage points in real time for very large scale maps. Due to restricting the search space of these vantage points through the winning region in a surveillance game, we are also able to guarantee correctness on these large-scale maps. 

We note that for illustration, the experiments shown here use a simple surveillance specification $\varphi = \square p_{1}$ which means the agent cannot lose sight of the target. We provide additional examples and animations with looser surveillance requirements at \url{http://visibility.page.link/evasion}. 

\section{Conclusion}
We proposed a framework for constraining patrol strategies to be formally correct with respect to a surveillance requirement defined as a temporal logic constraint. Our approach is flexible as it allows a user to tailor the surveillance requirement for their specific application. Using convolutional neural networks, we compute a measure of information gain which prioritizes observing states that have not been seen for a given time interval. We choose the next vantage point as the location within the maximal reach set $R^\ast$ that optimizes the gain function. We show that employing the patrol strategy leads to better visibility of the environment, as measured by a receding gain metric.

For future work, we will extend the framework to handle the case of \emph{exploration}, where the environment is initially unknown.
We also plan to consider more sophisticated surveillance requirements, such as \emph{surveillance liveness objectives}, where, infinitely often, the agent must keep the uncertainty of the target's location below a predefined threshold.


\section{Acknowledgment}
This work was partially supported by NSF grant DMS-1720171, AFRL FA9550-19-1-0169, DARPA D19AP00004, ONR N00014-18-1-2829. We thank Texas Advanced Computing Center (TACC) for providing the computational resources that made this work possible.

\bibliographystyle{IEEEtran}
\bibliography{main}

\end{document}